\documentclass[sigconf, nonacm]{acmart}
\AtBeginDocument{%
  \providecommand\BibTeX{{%
    \normalfont B\kern-0.5em{\scshape i\kern-0.25em b}\kern-0.8em\TeX}}}
\usepackage{soul}
\usepackage{enumitem}
\setcopyright{acmlicensed}
\copyrightyear{2018}
\acmYear{2018}
\acmDOI{XXXXXXX.XXXXXXX}

\acmConference[Conference acronym 'XX]{Make sure to enter the correct
  conference title from your rights confirmation emai}{June 03--05,
  2018}{Woodstock, NY}
\acmISBN{978-1-4503-XXXX-X/18/06}

\begin{document}

\title{ERR@HRI 2024 Challenge:  Multimodal  Detection of Errors and Failures in Human-Robot Interactions}

\author{Micol Spitale}
\authornote{The author is also affiliated with Politecnico di Milano, Milan, Italy}
\email{ms2871@cam.ac.uk}
\affiliation{%
  \institution{University of Cambridge}
  \streetaddress{XX}
  \city{Cambridge}
  \country{UK}
}

\author{Maria Teresa Parreira}
\affiliation{%
  \streetaddress{Bill and Melinda Gates Hall, 236}
  \institution{Cornell University}
  \city{Ithaca, NY}
  \country{USA}
}

\author{Maia Stiber}
\affiliation{%
  \streetaddress{Malone Hall, 3400 N Charles St}
  \institution{Johns Hopkins University}
  \city{Baltimore, MD}
  \country{USA}
}

\author{Minja Axelsson}
\affiliation{%
  \streetaddress{15 JJ Thomson Ave}
  \institution{University of Cambridge}
  \city{Cambridge}
  \country{UK}
}

\author{Neval Kara}
\authornote{Contributed to this work while undertaking a remote visiting studentship at
Department of Computer Science and Technology, University of Cambridge.}
\affiliation{%
  \institution{Cankaya University}
  \city{Ankara}
  \country{Turkey}
}

\author{Garima Kankariya}
\authornote{Contributed to this work while undertaking a remote visiting studentship at
Department of Computer Science and Technology, University of Cambridge.}
\affiliation{%
  \institution{Indian Institute of Technology}
  \city{Delhi}
  \country{India}
}

\author{Chien-Ming Huang }
\affiliation{%
  \streetaddress{Malone Hall, 3400 N Charles St}
  \institution{Johns Hopkins University}
  \city{Baltimore, MD}
  \country{USA}
}

\author{Malte Jung}
\affiliation{%
  \streetaddress{Bill and Melinda Gates Hall, 236}
  \institution{Cornell University}
  \city{Ithaca, NY}
  \country{USA}
}

\author{Wendy Ju}
\affiliation{%
  \streetaddress{2 W Loop Road}
  \institution{Cornell Tech}
  \city{New York, NY}
  \country{USA}
}

\author{Hatice Gunes}
\affiliation{%
  \streetaddress{15 JJ Thomson Ave}
  \institution{University of Cambridge}
  \city{Cambridge}
  \country{UK}
}

\renewcommand{\shortauthors}{Spitale et al.}
\renewcommand{\shorttitle}{ERR@HRI}

\begin{abstract}

Despite the recent advancements in robotics and machine learning (ML), the deployment of autonomous robots in our everyday lives is still an open challenge. This is due to multiple reasons among which are their frequent mistakes, such as interrupting people or having delayed responses, as well as their limited ability to understand human speech, i.e., failure in tasks like transcribing speech to text. These mistakes may disrupt interactions and negatively influence human perception of these robots. To address this problem, robots need to have the ability to detect human-robot interaction (HRI) failures. 
The ERR@HRI 2024 challenge tackles this by offering a benchmark multimodal dataset of robot failures during human-robot interactions (HRI), encouraging researchers to develop and benchmark multimodal machine learning models to detect these failures.
We created a dataset featuring multimodal non-verbal interaction data, including facial, speech, and pose features from video clips of interactions with a robotic coach, annotated with labels indicating the presence or absence of robot mistakes, user awkwardness, and interaction ruptures, allowing for the training and evaluation of predictive models. 
Challenge participants have been invited to submit their multimodal ML models for detection of robot errors and to be evaluated against various performance metrics such as accuracy, precision, recall, F1 score, with and without a margin of error reflecting the time-sensitivity of these metrics.
The results of this challenge will help the research field in better understanding the robot failures in human-robot interactions and designing autonomous robots that can mitigate their own errors after successfully detecting them.
\end{abstract}




\maketitle

\section{Introduction}
Human-Robot Interaction (HRI) research is currently placing a greater emphasis on the development of autonomous robots that can be deployed in real-world scenarios to understand the implications of integrating such robots in our lives. However, past works \cite{spitale2023longitudinal, parreira2023robot, stiber2023using} have shown that such autonomous robots are often characterised by making mistakes, for example when the robot interrupts people or when the robot takes a very long time to respond. These robot failures may disrupt the interaction and negatively impact the perception of people towards the robot \cite{spitale2023vita}. To overcome this problem, robots should be able to detect HRI failures.

The \emph{ERR@HRI 2024} challenge aims at addressing this issue by providing the community with a benchmark multimodal dataset of robot failures during human-robot interaction. The challenge encourages researchers to benchmark and develop multimodal machine learning-based models designed to identify when failures occur during HRI. 

We recruited challenge participants through email advertisements (e.g., ICMI announcements, robotics-worldwide) that included a link to our website\footnote{\url{https://sites.google.com/cam.ac.uk/err-hri/home}} where they could fill out the registration form. An EULA agreement, approved by both the DPO and the Departmental Ethics Committee of the University of Cambridge, was shared with the teams who signed up. The signed EULA was then sent to the research office of the University of Cambridge for a final review and approval.

We provided participants with a dataset that includes 1) multimodal non-verbal features (i.e., facial, speech, and pose features) of interaction clips where individuals interact with a robotic coach delivering positive psychology exercises, and 2) binary labels in the form of ‘interaction rupture present’ (1) or ‘interaction rupture absent’ (0). 
These features and labels were to be used to train the predictive models. The dataset was annotated as a time-series with the following labels: robot mistake (e.g., interruption or non-responding, (0) absent, (1) present), user awkwardness (e.g., when the participant feels uncomfortable interacting with the robot without any robot mistakes, (0) absent, (1) present), and interaction ruptures (i.e., either when the user displays some cues of awkwardness towards the robot and/or when the robot makes some mistakes;  (0) absent, (1) present). We invited the teams to submit their multimodal ML models  for error detection to be evaluated and benchmarked against the pre-determined performance metrics, including accuracy, precision, recall, F1 score, with and without an error margin \cite{parreira2023robot,de2012survey}.

\subsection{Relevance to multimodal interaction}
This challenge aims at addressing the problem of detecting robot failures in human-robot interaction, and as such it is extremely relevant to the multimodal interaction community. HRI is multimodal by nature because interactions often involve multiple types of social signaling, such as facial expressions, speech and body language of both humans and robots that, if better understood, can be used as cues to facilitate more natural interactions. ERR@HRI provides a novel multimodal dataset that can be used by participants to develop multimodal machine learning failure detection models. By highlighting the use of multimodal datasets and ML models for detecting failures, the ERR@HRI challenge contributes to advancing the understanding and enhancement of interactions between humans and autonomous robots in real-world settings.
The increased interest of the ICMI community in HRI is also evident by the recent contributions published in ICMI proceedings that include 5 papers at ICMI’23 (e.g., \cite{kalatzis2023multimodal}) and 2 at ICMI’22 (e.g., \cite{tan2022group}) on HRI, and as well as a keynote talk by Prof Maja Mataric at ICMI’23 entitled “A Robot Just for You: Multimodal Personalized Human-Robot Interaction and the Future of Work and Care”. The talk focused on multimodal aspects of HRI in healthcare, demonstrating the ICMI community’s increasing attention to this field.

\section{Related Work}

Past works have shown that robot failures are known to commonly occur during human-robot interactions, and they can negatively impact the user’s trust towards the robot. For example, Spitale et al. \cite{spitale2023robotic} demonstrated that participants experienced frustration when the robot interrupted them e.g., by erroneously detecting the end of the user's speech while they were still talking, or when the robot exhibited prolonged response times due to internet connectivity issues. Analogously, Kontogiorgos et al. examined human non-verbal behaviour reactions to conversational failures during a cooking instruction class delivered by a Furhat robot \cite{kontogiorgos2020behavioural, kontogiorgos2021systematic}. They found that severe errors may decrease users’ trust in the robot \cite{kontogiorgos2021systematic}. However, very few works attempted to address this problem by automatically detecting such failures. Spitale et al. \cite{spitale2023vita} introduced a new multimodal LLM-based system that allows robotic coaches to autonomously adapt to individual’s multimodal behaviours (facial valence and speech duration) and detect ruptures while delivering well-being coaching. Bremers et al. \cite{bremers2023bystander} used the bystander reaction dataset as input to a deep-learning model, BADNet, to predict failure occurrence without levering multimodal information. These studies represent the first stepping stone toward identifying robot failures during HRI, but they neither focused on benchmarking nor organising a challenge event to enable such comparisons under pre-defined settings and metrics.

The ERR@HRI initiative will provide a unique opportunity for benchmarking not only HRI data but also multimodal machine learning models to detect interaction ruptures,  which is fundamental for the success of human-robot interactions. In this first edition, the challenge will focus on using a multimodal dataset collected in a real-world setting where a robotic coach delivered well-being coaching practices to each participant over four weeks. For future editions of the challenge, we plan to focus on additional datasets, such as REACT and Response to Errors in HRI \cite{stiber2023using}, which have already been collected by the co-organizers of this challenge. This will enable a sustained engagement of the research community over the next couple of years and push the state of the art in multimodal robot failure analysis, detection and understanding.

\section{The ERR@HRI 2024 Challenge}

This section describes the dataset provided, including feature extraction, tasks, and evaluation process.

\subsection{Materials}
A challenge website was set up\footnote{\url{https://sites.google.com/cam.ac.uk/err-hri/home}} with a commitment to be maintained at least for the next 3 years. 
A GitHub repository\footnote{\url{https://github.com/ERR-HRI-Challenge/baseline2024}} has been established along with the official website to guide and support the challenge participants.

\subsection{Feature Extraction}

We used a dataset collected in a previous study \cite{spitale2023longitudinal, axelsson2024oh}, in which we deployed a robotic positive psychology coach at a workplace over four weeks. We involved a total of 43 employees out which 23 gave consent to share their data in processed and aggregated form. The robotic coach conducted four positive psychology exercises over four weeks. Please check the paper \cite{spitale2023robotic} for more detail on the study. 
During the interaction, we collected video recordings (coachee’s face and a side view of the interaction) and audio recordings (both the coachee’s and robot’s speech) using two cameras (a frontal video camera and a lateral GoPro) and a Jabra microphone. 

We used off-the-shelf state-of-the-art methods to extract multimodal behavioural features from the audio-visual data collected from the side-view camera as follows:
\begin{enumerate}
    \item Facial Features: We used the OpenFace 2.2.0 toolkit to extract the presence and intensity of 17 facial action units (AUs), in a total of 35 facial features per frame, at a rate of 30 fps.
    \item Audio Features: We used the openSMILE toolbox and extracted a total of 25 features, corresponding to the low level descriptors of feature set eGeMAPSv02, using a time window of 0.02 s and at a rate of 100 data points per second.  
    \item Pose Features: We used the OpenPose toolbox \cite{openpose} and extracted the 25-2D body key points per frame to estimate the movement of the torso, hands, arms, and head. The features provided (at 30 fps) do not correspond directly to the features extracted from Openpose, but rather the relational distance and velocity for pairs of spatial body points, in a total of 44 features, corresponding to relational features of  25 body points.

\end{enumerate}

\subsection{Labels}
The video clips were labelled by 2 annotators using the ELAN video annotation tool. We marked instances of user awkwardness and robot mistakes with binary labels (i.e., 1: present, or 0: absent), marking the time when the displays of user awkwardness or robot mistakes start and end. These labels have been defined in \cite{spitale2023longitudinal} as follows: 
\begin{itemize}
    \item \textbf{User Awkwardness (UA)}: The coachee displays behaviours that signal the interaction is awkward  — they may look confused, uncertain, distressed or uncomfortable. 
    \item \textbf{Robot Mistake (RM)}: The robot makes a mistake such as interrupting or not responding to the coachee, or responding with an utterance that is not appropriate for what the coachee has just said. 
    \item \textbf{Interaction Rupture (IR)}: We define an interaction rupture as either the presence of user awkwardness, a robot mistake, or both.

\end{itemize}

\subsection{Sub-challenges}
Accordingly, the ERR@HRI 2024 Challenge consists of the following three sub-challenges:
\begin{enumerate}
    \item Detection of robot mistakes (e.g., interrupting or not responding to the coachee)
    \item Detection of user awkwardness (e.g., when the coachee feels uncomfortable interacting with the robot without any robot mistakes)
    \item Detection of interaction ruptures (i.e. when the robot makes mistakes as described in 1) or when user displays awkwardness towards the robot described in 2))
\end{enumerate}

\subsection{Dataset}

The dataset contains data from 23 users, in a total of 89 sessions and 700 minutes of interaction. 

ERR@HRI 2024 participants are provided with 4 suggested dataset splits (i.e., subject-independent folds), with no overlapping participant data. Details of the data are provided in Table \ref{tab:dataset}.

\begin{table*}[]
    \centering
    \caption{\small Dataset and ground truth characteristics. Time per label includes the total amount of time within the dataset labeled as that type of interaction failure. Percentage refers to time per label over total time -- which provides a sense of dataset label balance.}

    \begin{tabular}{l|lllllllll}

\textbf{Subset}
    & \textbf{Subjects} 
    & \textbf{Sessions}
    & \textbf{Total time (s)}
    & \textbf{Time RM (s)} 
    & \textbf{\% RM}
    & \textbf{Time UA (s)} 
    & \textbf{\% UA} 
    & \textbf{Time IR (s)} 
    & \textbf{\% IR}
    \\ \hline
\textbf{Train + Val}    & 18 
    & 71 
    & 33308 
    & 5320 
    & 0.16 
    & 5182 
    & 0.16 
    & 8679 
    & 0.26 
    \\ 
        \textbf{Test} 
    & 5 
    & 18 
    & 8048 
    & 1399 
    & 0.17 
    & 1875 
    & 0.23 
    & 2738 
    & 0.34 
    \end{tabular}
    \label{tab:dataset}
\end{table*}

\subsection{Metrics}

This challenge contemplates three binary classification tasks. The metrics used to evaluate model performance are accuracy, precision, recall, f1-score, as well as metrics with a margin of error \cite{parreira2023robot,de2012survey} -- for a sample margin of size $k$, and for a sample $i$, the model prediction is considered right if $y^i_{pred} \in [y^{i-k}_{pred},y^{i+k}_{pred}]$

The motivation for considering metrics with a margin of error is due to considerations of real-life settings where effectiveness may still be achieved even if the model is slightly early or delayed in its error detection. Other options for real-use systems could be to use the median or mode of predictions within an interval, among others. Metrics with a margin of error, in this challenge, include accuracy, precision, recall and f1-score.

\subsection{Evaluation}
Challenge participants were given access to the training and validation sets to develop their ML models. Then, they were asked to submit the developed models and weights, and the organisers have evaluated the submitted models on the test set (the test set was released to the challenge participants without labels one week prior to the submission deadline). Each participating group was allowed to submit their models and results for the test set up to three times. 
The submitted models and predictions were automatically evaluated and ranked using various performance metrics, under two categories: overall performance and marginal performance. For both tracks, models are ranked based on the combined rankings of accuracy and F1-score (for the marginal track, we use the accuracy and F1-score considering an error margin of 1 sample).
Metrics were calculated using the same script provided to participants in the study repository. Challenge participants were also asked to submit a paper describing their model via the EasyChair system, and their works were reviewed by the Technical Program Committee members of the challenge.


\section{Challenge Baseline}
We have provided a deep-learning multimodal baseline for each of the three tasks, as in \cite{bremers2023bystander} and \cite{spitale2023vita} (where we reported results for interaction rupture prediction).

\subsection{Training}

For baseline models, and following previous work on a similar dataset, we decided to use Recurrent Neural Network models, which can conserve some feature history and are common approaches for time-series classification problems in HRI. Namely, we made use of Long Short-Term Memory networks (LSTMs), Bidirectional-LSTMs (BiLSTMs), Gated Recurrent Unit networks (GRUs), which tend to overfit less than LSTMs in smaller datasets. We used single-layer models, with dropout and a fully-connected layer. We wanted to provide a standard approach to model development, leaving room for participants to innovate their approaches for detection and classification. 
For training, we did hyperparameter tuning using test accuracy as the metric to pick the top performing hyperparameters. We used a 3-1 train-validation fold split, with the suggested folds provided in the study repository. For each task and each model architecture, we picked the top 3 performing model hyperparameters.-- a total of 9 models per task. These models were then trained using cross-validation on the 4 folds and the final model was picked based on the average metrics across all folds. In the end, each of these models was trained on the 4 folds and predictions on the test set were reported to all participants. 
\begin{table*}[htb!]

\centering
\small
\caption{\small Hyperparameters of best performing models. SL: sequence length. LR: learning rate.}
\label{tab:models}

\begin{tabular}{lcc}

\multicolumn{1}{c}{\textbf{Task}}  & \multicolumn{1}{c}{\textbf{Model}}   & \multicolumn{1}{c}{\textbf{Hyperparameters}} \\ \hline

\textit{RM} & GRU & \multicolumn{1}{c}{\begin{tabular}[c]{@{}c@{}} SL=5, Units=128, Dropout=0.2, LR: 0.0001 \\ Activation: softmax, Optimizer: Adam \\ Loss: Categorical Cross-Entropy, Batch size = 2048, Epochs = 500 \end{tabular}}  \\ \hline

\textit{UA} & BiLSTM & \multicolumn{1}{c}{\begin{tabular}[c]{@{}c@{}} SL=5, Units=256, Dropout=0.2, LR: 0.0001 \\ Activation: sigmoid, Optimizer: Adam \\ Loss: Categorical Cross-Entropy, Batch size = 512, Epochs = 200 \end{tabular}}  \\ \hline

\textit{IR}  & BiLSTM & \multicolumn{1}{c}{\begin{tabular}[c]{@{}c@{}} SL=5, Units=256, Dropout=0.2, LR: 0.0001 \\ Activation: softmax, Optimizer: Adam \\ Loss: Categorical Cross-Entropy, Batch size = 4096, Epochs = 500 \end{tabular}}  \\ 

\end{tabular}
\end{table*}

\begin{table*}[htb!]
\vspace{4pt}
\centering
\small
\caption{\small Baseline (macro) performances. Margin of error metrics are noted with and $e$ and represent a 1-sample tolerance.}
\label{tab:performance}

\begin{tabular}{lcccccccc}

\textbf{Task}  & \textbf{Accuracy} & \textbf{Precision} & \textbf{Recall} & \textbf{F1-Score}& \textbf{Accuracy$_e$} & \textbf{Precision$_e$} & \textbf{Recall$_e$} & \textbf{F1-Score$_e$}\\ \hline

\textit{RM} & $0.71349$ & $0.55593$ & $0.54089$ & $0.54184$ & $0.71417$ & $0.55756$ & $0.54219$ & $0.54335$  \\ 

\textit{UA} & $0.73074$ & $0.56358$ & $0.57356$ & $0.56698$ & $0.73207$ & $0.56617$ & $0.57676$ & $0.56978$ \\ 

\textit{IR} & $0.68460$ & $0.55541$ & $0.50268$ & $0.41964$ & $0.68592$ & $0.58794$ & $0.50478$ & $0.42395$ \\ 

\end{tabular}
\end{table*}

\subsection{Results} The hyperparameters and performance for each model, for each task, are described in Tables \ref{tab:models} and \ref{tab:performance}. The best performing models have short sequence lengths (5 samples) and the BiLSTM  model performed best across two of the subchallenges. The obtained perfomances on the test set are slightly above chance level.

\section{Participation and Conclusion}
This paper introduced the ERR@HRI 2024 Challenge organised in conjunction with the ACM International Conference on Multimodal Interaction 2024 (ACM ICMI'24), which focuses on detecting robot failures in human-robot interactions. 
A total of 10 teams from 5 countries registered for this challenge, and 3 teams submitted their results for benchmarking and evaluation. 
The submitted models will be ranked under identical conditions using the specified evaluation protocol and metrics. We aim for the challenge data, code, systems, and results from competing teams to be valuable resources for researchers and practitioners focused on detecting failures in human-robot interactions.
Our future efforts will be directed at continuing to organize ERR@HRI challenge events in conjunction with well-known conferences while introducing new datasets and modalities.


\begin{acks}
\noindent\textbf{Funding:} This challenge is possible due to the EPSRC/UKRI grant EP/R030782/1 (ARoEQ) and EP/R511675/1 that supported the HRI studies, and the work of M. Spitale and H. Gunes, that generated the data used in this challenge. 
M. Spitale's current work involving the organisation of this challenge and the writing of this paper is supported by PNRR-PE-AI FAIR project funded by the NextGeneration EU program. \\
\textbf{Open Access:} For open access purposes, the authors have applied a Creative Commons Attribution (CC BY) licence to any Author Accepted Manuscript version arising.\\
\textbf{Data access:} Raw data related to this publication cannot be openly released due to anonymity and privacy issues. However, challenge participants who signed the EULA agreement have been granted access to processed data in the form of aggregated feature statistics and models. 
\end{acks}

\bibliographystyle{ACM-Reference-Format}
\bibliography{ref}


\end{document}